%% file: main.tex
\documentclass[letterpaper, 10 pt, conference]{ieeeconf}
\IEEEoverridecommandlockouts
\let\labelindent\relax
\usepackage{enumitem}
\usepackage{balance}
\usepackage[font={small}]{caption}
\usepackage{subcaption}
\usepackage{array}
\usepackage{textcomp}
\usepackage{mathtools, nccmath}
\usepackage{graphicx}
\usepackage{amsfonts}
\usepackage{amsmath}
\usepackage{amssymb}
\usepackage{algorithm}
\usepackage{algorithmic}
\usepackage{hyperref}
\usepackage{tikz}
\usepackage{arydshln}
\usepackage{multirow}
\usepackage{bm}
\usepackage{epstopdf}
\usepackage{cite}
\usepackage{siunitx}
\usepackage{booktabs}

\DeclareMathOperator*{\argmin}{argmin} 

\usepackage{amsthm}

\newtheoremstyle{mystyle}
  {}
  {}
  {}
  {}
  {\bfseries}
  {.}
  { }
  {\thmname{#1}\thmnumber{ #2}\thmnote{ (#3)}}

\theoremstyle{mystyle}

\newtheorem{remark}{Remark}

\title{\LARGE \bf Autonomous Racing with Multiple Vehicles using a Parallelized Optimization with Safety Guarantee using Control Barrier Functions}
\begin{document}
\author{Suiyi He$^{1*}$, Jun Zeng$^{2*}$, and Koushil Sreenath$^2$
\thanks{
$^*$Authors have contributed equally and listed alphabetically.}
\thanks{$^1$Author is with the Department of Mechanical Engineering, University of Minnesota-Twin Cities, MN 55455. {\tt\small he000231@umn.edu}}
\thanks{$^2$Authors are with the Hybrid Robotics Group at the Department of Mechanical Engineering, University of California, Berkeley. {\tt\small \{zengjunsjtu,koushils\}@berkeley.edu}}
\thanks{This work was partially supported through National Science Foundation Grant CMMI-1931853.}
\thanks{The animation video can be found at \url{https://youtu.be/1zTXfzdQ8w4} and the implementation code can be found on \url{https://github.com/HybridRobotics/car-racing}.}
}
\maketitle

\begin{abstract}
This paper presents a novel planning and control strategy for competing with multiple vehicles in a car racing scenario.
The proposed racing strategy switches between two modes.
When there are no surrounding vehicles, a learning-based model predictive control (MPC) trajectory planner is used to guarantee that the ego vehicle achieves better lap timing performance.
When the ego vehicle is competing with other surrounding vehicles to overtake, an optimization-based planner generates multiple dynamically-feasible trajectories through parallel computation. Each trajectory is optimized under a MPC formulation with different homotopic Bezier-curve reference paths lying laterally between surrounding vehicles.
The time-optimal trajectory among these different homotopic trajectories is selected and a low-level MPC controller with control barrier function constraints for obstacle avoidance is used to guarantee system's safety-critical performance.
The proposed algorithm has the capability to generate collision-free trajectories and track them while enhancing the lap timing performance with steady low computational complexity, outperforming existing approaches in both timing and performance for a autonomous racing environment.
To demonstrate the performance of our racing strategy, we simulate with multiple randomly generated moving vehicles on the track and test the ego vehicle's overtake maneuvers.
\end{abstract}

\IEEEpeerreviewmaketitle
\input{sections/introduction}
\input{sections/background}
\input{sections/racing_algorithm}

\input{sections/results}
\input{sections/conclusion}

\balance
\bibliographystyle{IEEEtran}
\typeout{}\bibliography{reference}{}
\end{document}

%% file: sections/introduction.tex
\section{Introduction}
\label{sec:introduction}
\subsection{Motivation}
\begin{figure}[t]
    \setlength{\abovecaptionskip}{0.5cm}
    \setlength{\belowcaptionskip}{-0.5cm}
    \centering
    \includegraphics[width=0.8\linewidth]{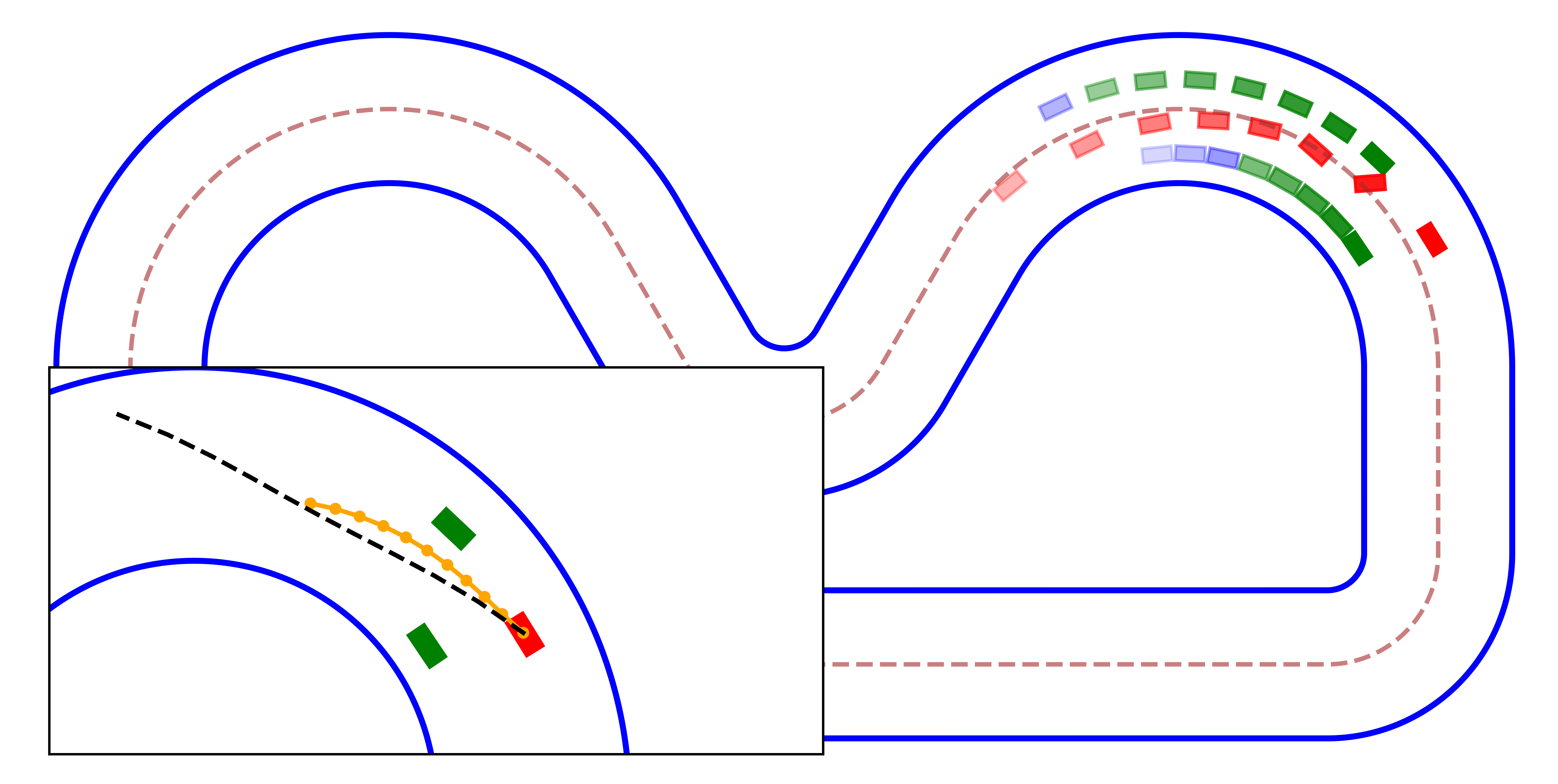}
    \caption{Snapshots from simulation of the overtaking behavior. Red is the ego vehicle and the color of the surrounding vehicles switches from blue to green depending on if they are in the ego vehicle's range of overtaking. Note the large sideways orientation of the red ego car at the start of the overtake maneuver.
    The solid blue line and dashed brown line are track's boundary and center line. The dashed black line and yellow line (in inset) are  reference paths and the optimized trajectory, respectively.}
    \label{fig:introduction_snapshot}
\end{figure}

Recently, autonomous racing is an active subtopic in the field of autonomous driving research.
In autonomous racing, the ego car is required to drive along a specific track with an aggressive behavior, such that it is capable of competing with other agents on the same track.
By overtaking other leading vehicles and moving ahead, the ego vehicle can finish the racing competition with a smaller lap time.
While the behavior of overtaking other vehicles has been studied in autonomous driving on public roads, however, these techniques are not effective on a race track.
This is because autonomous vehicles are guided by dedicated lanes on public roads to succeed in lane follow and lane change behaviors, while the racing vehicles compete in the limited-width tracks without guidance from well-defined lanes.
Existing work focuses on a variety of algorithms for autonomous racing, but most of them could not provide a time-optimal behavior with high update frequency in the presence of other moving agents on the race track.
In order to generate racing behaviors for the ego racing car, we propose a racing algorithm for planning and control that enables the ego vehicle to maintain time-optimal maneuvers in the absence of local vehicles, and fast overtake maneuvers when local vehicles exist, as shown in Fig.~\ref{fig:introduction_snapshot}. 

\subsection{Related Work}
In recent years, researchers have been focusing on planning and control for autonomous driving on public roads. For competitive scenarios like autonomous lane change or lane merge, both model-based methods \cite{schmidt2019interaction} and learning-based methods \cite{yao2012learning} have been demonstrated to generate the ego vehicle's desired trajectory.
Similarly, control using model-based methods \cite{vazquez2020optimization, he2021cbf, lyu2021probabilistic} and learning-based methods \cite{krasowski2020safe} has also been developed.
However, the criteria to evaluate planning and control performance are different for car racing compared to autonomous driving on public roads.
For autonomous racing~\cite{loiacono2013simulated}, when the ego racing car competes with other surrounding vehicles, most on-road traffic rules are not effective.
Instead of maneuvers that offer a smooth and safe ride, aggressive maneuvers that push the vehicle to its dynamics limit \cite{wischnewski2019model, talvala2011pushing} or even beyond its dynamics limit \cite{goh2020toward} are sought to win the race.
In order to quickly overtake surrounding vehicles, overtake maneuvers with tiny distances between the cars and large orientation changes are needed.
Moreover, due to the bigger slip angle caused by changing the steering orientation more quickly during racing, more accurate dynamical models should be used for autonomous racing planning and control design.  We next enumerate the related work in several specific areas.

\begin{table*}
\centering
\scriptsize
\begin{tabular}{|c | c| c |c |c |c |c |c |c |c |c |c |c |c|}
\hline
\emph{Approach} & \emph{GP} & \emph{DRL} & \emph{Graph-Search} & \multicolumn{2}{c|}{\emph{Game Theory}} & \multicolumn{8}{c|}{\emph{Model-Based}} \\ \hline
Publication & \cite{hewing2018cautious} & \cite{fuchs2021super} & \cite{stahl2019multilayer} & \cite{liniger2019noncooperative} & \cite{wang2021game} & \cite{heilmeier2019minimum} & \cite{rosolia2017learning} & \cite{kabzan2019learning} & \cite{kapania2020learning} & \cite{febbo2017moving} & \cite{brudigam2021gaussian} & \cite{zeng2021safety} & \textbf{Ours} \\ \hline
Lap Timing & \textbf{Yes} & \textbf{Yes} & \textbf{Yes} & \textbf{Yes} & No & \textbf{Yes} & \textbf{Yes} & \textbf{Yes} & \textbf{Yes} & No & No & No & \textbf{Yes} \\ \hline
Static Agent & No & No & \textbf{Yes} & \textbf{Yes} & \textbf{Yes} & No & No & No & No & \textbf{Yes} & \textbf{Yes} & \textbf{Yes} & \textbf{Yes} \\ \hline
Moving Agent & No & No & One & One & \textbf{Multiple} & No & No & No & No & \textbf{Multiple} & One & One & \textbf{Multiple} \\ \hline
Update Frequency (Hz) & N/A & N/A & 15 & \textbf{30} & 2 & Offline & \textbf{20} & \textbf{20} & Offline & \textless 1 & 10 & \textless 10 & \textgreater\textbf{25} \\ \hline
Planner & No & No & \textbf{Yes} & \textbf{Yes} & \textbf{Yes} & \textbf{Yes} & No & No & No & \textbf{Yes} & \textbf{Yes} & No & \textbf{Yes} \\ \hline
Dynamics Accuracy & N/A & N/A & \textbf{Yes} & \textbf{Yes} & No & \textbf{Yes} & \textbf{Yes} & \textbf{Yes} & \textbf{Yes} & \textbf{Yes} & \textbf{Yes} & \textbf{Yes} & \textbf{Yes} \\ \hline
\end{tabular}
\normalsize
\caption{A comparison of recent work on autonomous racing and their attributes. Lap timing indicates if the lap timing performance is considered. Static and moving agent indicates if other static or moving agents are considered. Update frequency indicates the optimization update frequency. Learning-based approaches like GP don't have this attribute. Planner indicates if the approach has the planning part. Dynamics accuracy indicates the dynamics model used in the controller, with ``Yes", ``No", ``N/A" representing dynamic model, kinematic model and model-free.
}
\label{tab:work_summary}
\end{table*}

\subsubsection{Planning Algorithms}
For autonomous racing, the planner is desired to generate a time-optimal trajectory.
Although some work using convex optimization problems \cite{liniger2015optimization, kapania2016sequential, bonab2019optimization, caporale2019towards, heilmeier2019minimum, srinivasan2021holistic, alcala2020autonomous} or Bayesian optimization (BO) \cite{jain2020computing} reduces the ego vehicle's lap time impressively, either no obstacles \cite{kapania2016sequential, bonab2019optimization, caporale2019towards, heilmeier2019minimum, srinivasan2021holistic, alcala2020autonomous, jain2020computing} or only static obstacles \cite{liniger2015optimization} are assumed to be on the track.    
When moving vehicles exist on the track, nonlinear dynamic programming (NLP) \cite{febbo2017moving}, graph-search \cite{stahl2019multilayer} and game theory \cite{liniger2019noncooperative, wang2021game, jung2021game} based approaches have demonstrated their capabilities to generate collision-free trajectories. 
Additionally, in order to improve the chance of overtaking, offline policies are learnt for the overtake maneuvers at different portions of a specific track \cite{bhargav2021track}.
However, these approaches don't solve all challenges.
For instance, work in \cite{febbo2017moving, wang2021game, jung2021game, bhargav2021track} does not take lap timing enhancement into account.
In \cite{stahl2019multilayer}, the ego vehicle is assumed to compete on a straight track with one constant-speed surrounding vehicle.
These assumptions are relatively simple for a real car racing competition.
In \cite{liniger2019noncooperative}, it is assumed that the planner knows the other vehicle's strategy and the complexity of the planner increases excessively when multiple vehicles compete with each other on the track.

\subsubsection{Control Algorithms}
Researchers focus on enhancing performance of the ego vehicle by achieving its speed and steering limits through better control design, e.g., obtaining the optimal lap time by driving fast.
The majority of existing work focuses on developing controllers with no other vehicles on the track.
The learning-based controllers \cite{rosolia2017learning,  kabzan2019learning, kapania2020learning, wang2021deep} leverage the control input bounds to achieve optimal performance in iterative tasks.
Model-free methods like Bayesian optimization (BO) \cite{oliveira2018learning}, Gaussian processes (GPs) \cite{hewing2018cautious}, deep neural networks (DNN) \cite{perot2017end, wadekar2021towards} and deep reinforcement learning (DRL) \cite{remonda2019formula, fuchs2021super} have also been exploited to develop controllers that result in agile maneuvers for the ego car.
To deal with other surrounding vehicles, DRL has also been used in \cite{song2021autonomous} to control the ego vehicle during overtake maneuvers.
Recently, model predictive based controllers (MPC) with nonlinear obstacle avoidance constraints have become popular to help the ego vehicle avoid other vehicles in the free space.
A nonconvex nonlinear optimization based controller is implemented in \cite{rosolia2016autonomous} to help the ego vehicle avoid static obstacles. 
Researchers in \cite{li2021autonomous} use mixed-integer quadratic programs (MIQP) to help the ego vehicle compete with one moving vehicle.
In \cite{brudigam2021gaussian}, GPs was applied to formulate the distance constraints of a stochastic MPC controller with a kinematic bicycle model.
However, large slip angles under aggressive maneuvers will cause a mismatch between real dynamics model and the kinematic model used in the controller, resulting in the controller being unable to guarantee the system's safety in some cases.
In \cite{zeng2021safety}, a safety-critical control design by using control barrier functions is proposed to generate a collision-free trajectory without a high-level planner, where infeasibility could arise due to the high nonlinearity of the optimization problem.
Moreover, due to the lack of a trajectory planner, deadlock could happen very often during overtake maneuvers, such as in \cite{zeng2021safety, brudigam2021gaussian}.  A comparison of various approaches and their features are enumerated in TABLE~\ref{tab:work_summary}.

As mentioned above, all the previous work on planning and control design for autonomous racing could not enhance the lap timing performance and simultaneously compete with multiple vehicles. Inspired by the work on iterative learning-based control and optimization-based planning, we propose a novel racing strategy to resolve the challenges mentioned above with a steady low computational complexity.

\subsection{Contribution}
The contributions of this paper are as follows:
\begin{itemize}
    \item We present an autonomous racing strategy that  switches between a learning-based MPC trajectory planner (in the absence of surrounding vehicles) and optimization-based homotopic trajectory planner with a low-level safety-critical controller (when the ego vehicle competes with surrounding vehicles).
    \item The learning-based MPC approach guarantees time-optimal performance in the absence of surrounding vehicles. When the ego vehicle competes with surrounding vehicles, multiple homotopic trajectories are optimized in parallel with different geometric reference paths and the best time-optimal trajectory is selected to be tracked with an optimization-based controller with obstacle avoidance constraints.
    \item We validate the robust performance together with steady low computational complexity of our racing strategy in numerical simulations where randomly moving vehicles are generated on a simulated race track. It is shown that our proposed strategy allows the ego vehicle to succeed in overtaking tasks without deadlock when there are multiple vehicles moving around the ego vehicle. We also demonstrate that our strategy would work for various racing environments.
\end{itemize}

%% file: sections/background.tex
\begin{figure*}
    \centering
    \includegraphics[width=\linewidth]{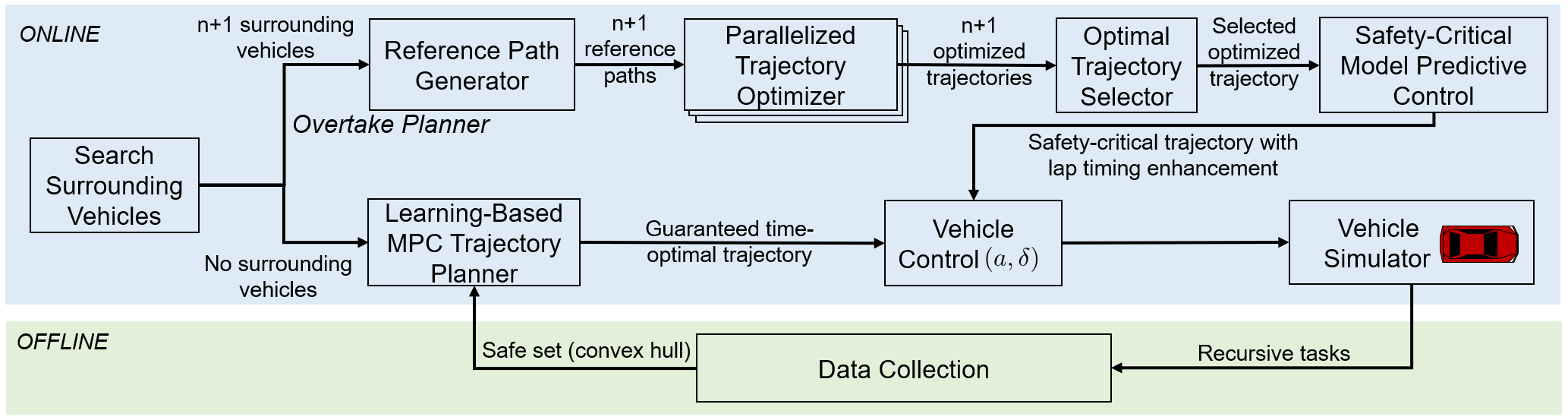}
    \caption{Autonomous Racing Strategy. The system dynamics is identified through offline data collection via recursive tasks. For online deployment, when no surrounding vehicles exist, the learning-based MPC trajectory planner is executed to guarantee time-optimal trajectories. When there are surrounding vehicles, the best time-optimal trajectory is chosen among the $n+1$ trajectories that are optimized in parallel with each optimization carried out for a particular homotopic trajectory around the $n$ surrounding cars. The chosen trajectory is then tracked with a safety-critical model predictive based controller.}
    \label{fig:pipeline}
\end{figure*}

\section{Background}\label{sec:background}
In this section, we revisit the vehicle model and learning-based MPC for iterative tasks. The learning-based MPC will be used as the trajectory planner when no surrounding vehicles exist.
\subsection{Vehicle Model}
In this work, we use a dynamic bicycle model with decoupled Pacejka tire model under Frenet coordinates. The system dynamics is described as follows,
\begin{equation}
\Dot{x} = f(x, u),
\label{eq:dynamic-model}
\end{equation}
where $x$ and $u$ show the state and input of the vehicle, $f$ is a nonlinear dynamic bicycle model in~\cite{rajamani2011vehicle}. The definition of state and input is as follows,
\begin{equation}
x=[v_x, v_y, \omega_z, e_\psi, s_c, e_y]^T,~u=[a, \delta]^T,
\end{equation}
where acceleration at vehicle's center of gravity $a$ and steering angle $\delta$ are the system's inputs.
$s_c$ denotes the curvilinear distance travelled along the track's center line, $e_y$ and $e_\psi$ show the deviation distance and heading angle error between vehicle and center line.
$v_x$, $v_y$ and $\omega_z$ are the longitudinal velocity, lateral velocity and yaw rate, respectively. 

In this paper, this model \eqref{eq:dynamic-model} is applied for precise numerical simulation using Euler discretization with sampling time 0.001s (1000Hz).
Through linear regression from the simulated reference path, an affine time-invariant model as below,
\begin{equation}
x_{t+1} = A(\bar{x}) x_{t} + B(\bar{x}) u_{t}
\label{eq:time_invariant_model}
\end{equation}
will be used in the trajectory planner to avoid excessive complexity from nonlinear optimization, where $\bar{x}$ represents the equilibrium point for linearized dynamics. 
On the other hand, an affine time-varying model as below,
\begin{equation}
x_{t+1} = A_{t}(\bar{x}_{k}) x_{t} + B_{t}(\bar{x}_{k}) u_{t} + C_{t}(\bar{x}_{k})
\label{eq:time_varying_model}
\end{equation}
where matrices $A_{t}(\bar{x}_{k})$, $B_{t}(\bar{x}_{k})$, and $C_{t}(\bar{x}_{k})$ are obtained at local equilibrium point $\bar{x}_{k}$ on reference trajectory with iterative data which is close to $x_t$.
The dynamics~\eqref{eq:time_varying_model} will be used on racing controller design for better tracking performance.

\subsection{Iterative Learning Control}
\label{subsec:lmpc}
A learning-based MPC \cite{rosolia2017learning}, which improves the ego vehicle's lap timing performance through iterative tasks, will be used in this paper.  This has the following components:
\subsubsection{Data Collection}
The learning-based MPC optimizes the lap timing through historical states and inputs from iterative tasks.
To collect initial data, a simple tracking controller like PID or MPC can be used for the first several laps.
During the data collection process, after the $j$-th iteration (lap), the controller will store the ego vehicle's closed-loop states and inputs as vectors.
Meanwhile, through offline calculation, every point of this iteration will be associated with a cost, which describes the time to finish the lap from this point. 
\subsubsection{Online Optimization}
\begin{figure}
    \setlength{\abovecaptionskip}{0.5cm}
    \setlength{\belowcaptionskip}{-0.5cm}
    \centering
    \includegraphics[width=1.0\linewidth]{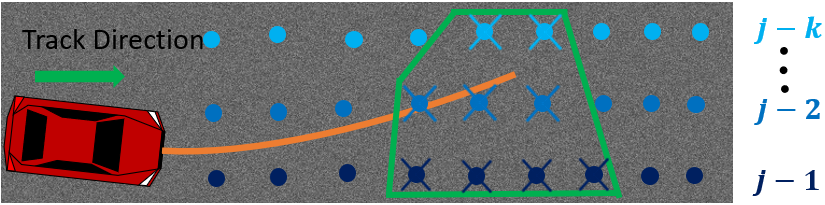}
    \caption{An illustration of learning-based MPC. Different groups of blue points show the historical closed-loop trajectories for different laps. When the ego vehicle is on the $j$-th lap, data from ($j-k$)-th to ($j-1$)-th lap can be used for online optimization. The figure illustrates this with $k=3$. Points with crosses are the selected neighboring historical states to the current point $x$. The green set is a convex approximation hull of these selected points, which is the terminal constraint for the MPC problem.
    The orange line is the open-loop optimized trajectory for the racing with $j$-th lap.
    This manner of iterative calculation generates time-optimal behavior when there are no surrounding vehicles.}
    \label{fig:lmpc_illustration}
\end{figure}

After the initial laps, the learning-based MPC optimizes the vehicle's behavior based on collected data.
At each time step, the terminal constraint is formulated as a convex set (green convex hull in Fig.~\ref{fig:lmpc_illustration}). This convex set includes the states that can drive the ego vehicle to the finish line in the previous laps.
By constructing the cost function to create a minimum-time problem, an open-loop optimized trajectory can be generated.
Since the cost function is based on the previous states' timing data, the vehicle is able to drive to the finish line with time that is no greater than the time from the same position during previous laps.
As a result, the ego vehicle will reach the time-optimal performance after several laps. 

More details of this method can be found in \cite{rosolia2017learning}. 
In our work, this approach will be used for trajectory planning when the ego vehicle has no surrounding vehicles. This helps with better lap timing without surrounding vehicles. Notice that the data for iterative learning control will be collected through offline simulation with no obstacles on the track, as shown in Fig.~\ref{fig:pipeline}.

%% file: sections/racing_algorithm.tex
\section{Racing Algorithm}
\label{sec:racing_algorithm}
After introducing the background of vehicle modeling and learning-based MPC, we will present an autonomous racing strategy that can help the ego vehicle enhance lap timing performance while overtaking other moving vehicles.

\subsection{Autonomous Racing Strategy}
\label{subsec:racing-strategy}
There are two tasks in autonomous racing: enhancing the lap timing performance and competing with other vehicles. 
To deal these two problems, our proposed strategy will switch between two different planning strategies.
When there are no surrounding vehicles, trajectory planning with learning-based MPC is used to enhance the timing performance through historical data.
Once the leading vehicles are close enough, an optimization-based trajectory planner optimizes several homotopic trajectories in parallel and the collision-free optimal trajectory is selected with an optimal-time criteria, which will be tracked by a low-level MPC controller.
By adding obstacle avoidance constraints to the low-level controller, it has the ability to guarantee the system's safety.
The racing strategy is summarized in Fig.~\ref{fig:pipeline}.

\subsection{Overtaking Planner}
To determine if a surrounding vehicle is in the ego vehicle's range of overtaking, following condition must be satisfied:
\begin{equation}
-\epsilon l \leq s_{c, i} - s_{c} \leq \epsilon l + \gamma |v_{x} - v_{x, i}|
\label{eq:overtaking_criteria}
\end{equation}
where $s_{c}$ and $s_{c, i}$ are ego vehicle's and $i$-th surrounding vehicle's traveling distance, $v_{x}$ and $v_{x, i}$ are ego vehicle's and $i$-th surrounding vehicle's longitudinal speed. $l$ indicates the vehicle's length. $\epsilon$ and $\gamma$ are safety-margin factor and prediction factor which we can tune for different performance.

\begin{figure}
    \centering
    \includegraphics[width=1.0\linewidth]{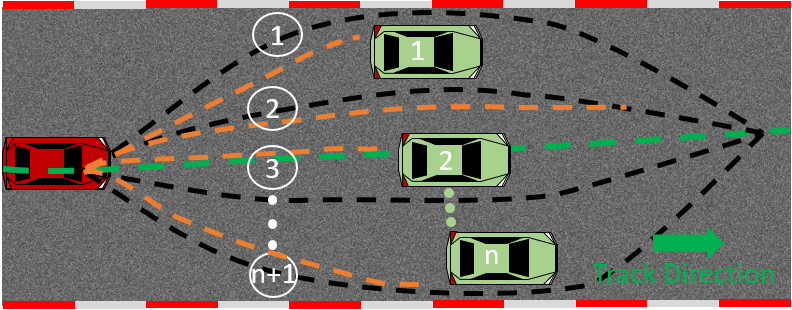}
    \caption{A typical overtaking scenario when there are $n$ vehicles in the range of overtaking.
    The ego vehicle and surrounding vehicles are in red and green, respectively. 
    The dashed green line is a time-optimal trajectory calculated from the learning-based MPC.
    Blue points are control points for Bezier-curves in two different cases.
    Dashed black lines are $n+1$ groups of reference paths each with a different homotopy.
    Dashed orange lines are optimized trajectories for each optimization problem.
    The trajectory in area 2 is selected as the best for its smoothness and reachability along the track.
    }
    \label{fig:overtaking-illustration}
\end{figure}

As shown in Fig.~\ref{fig:overtaking-illustration}, when there are $n$ vehicles in the ego vehicle's range of overtaking, there exists $(n+1)$ potential areas, each leading to paths with a different homotopy, that the ego vehicle can use to overtake these surrounding vehicles.
These areas are the one below the $n$-th vehicle, the one above the 1st vehicle, and the ones between each group of adjacent vehicles.
$n$+1 groups of optimization-based trajectory planning problems are solved in parallel, enabling steady low computational complexity even when competing with different numbers of surrounding vehicles.
To reduce each optimization problem's computational complexity through fast convergence, geometric paths with a distinct homotopy class that laterally lay between vehicles or vehicle and track boundary (black dashed curves in Fig.~\ref{fig:overtaking-illustration}) are used as reference paths in the optimization problems.
By comparing the optimization problems' costs, the optimal trajectory is selected from $n+1$ optimized solutions.
For example, as the case shown in Fig.~\ref{fig:overtaking-illustration}, the dashed orange line in area 2 will be selected since it avoids surrounding vehicles and finishes overtake maneuver with smaller time.
The function to minimize during the selection is shown as follows,
\begin{equation}
\begin{aligned}
&J_{s}(x_t) = \min_{x_{t}} -K_s (s_{c_{t+N}}-s_{c_t}) - \sum\limits_{k=1}^{N_p}  ((s_{c_{t+k}}\\&-s_{{c,i}_{t+k}})^2+(e_{y_{t+k}}-e_{{y,i}_{t+k}})^2-l^2-d^2) + b
\end{aligned}
\label{eq:cost-selection}
\end{equation}
where $K_s$ is a scalar used in metric for timing and $b$ is a non-zero penalty cost if the new potential area of overtaking is different from the area of overtaking in the last time step. 
A bigger value of $K_s$ is applied such that the ego vehicle is optimized to reach a farther point during the overtake maneuver, which results in a shorter overtaking time since the planner's prediction horizon and sampling time are fixed. Additionally, the other terms in \eqref{eq:cost-selection} prevents the ego vehicle from changing direction abruptly during an overtake maneuver and guarantees the ego vehicle's safety.

\begin{figure}
    \centering
    \begin{subfigure}[b]{\linewidth}
    \centering
    \includegraphics[width=\linewidth]{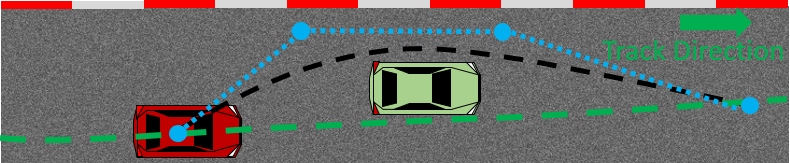}
    \caption{Control points for the Bezier-curve next to the track boundary.}
    \label{subfig:control-point-boundary}
    \end{subfigure}
    \begin{subfigure}[t]{\linewidth}
    \centering
    \includegraphics[width=\linewidth]{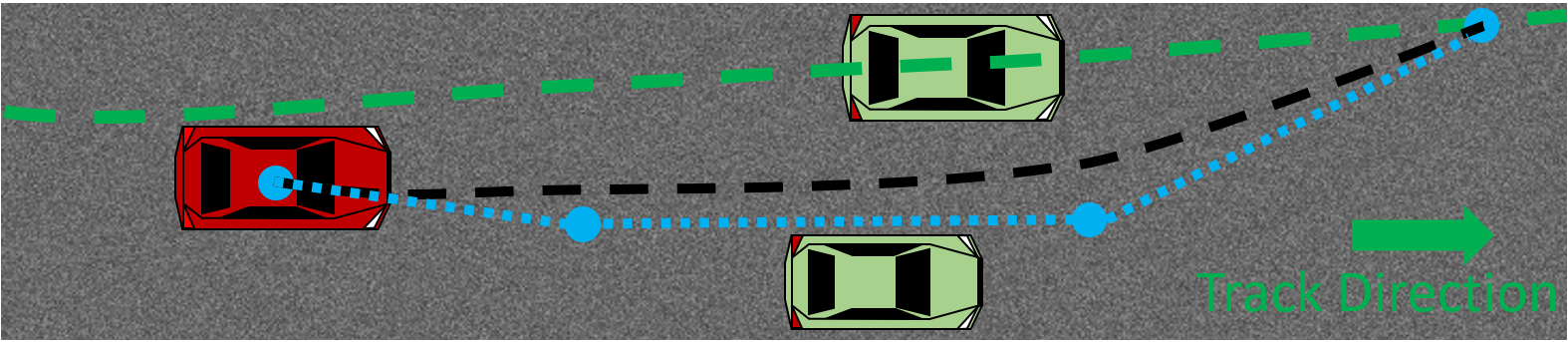}
    \caption{Control points for the Bezier-curve between two adjacent vehicles.}
    \label{subfig:control-point-vehicle}
    \end{subfigure}
    \caption{In the optimization problem, third order Bezier-curves are used as the reference paths. These two pictures show the Bezier-curves (dashed black lines) and their control points (blue points) for two different cases.}
    \label{fig:control-point}
\end{figure}

\begin{figure}
    \setlength{\abovecaptionskip}{0.5cm}
    \setlength{\belowcaptionskip}{-0.5cm}
    \centering
    \includegraphics[width=1.0\linewidth]{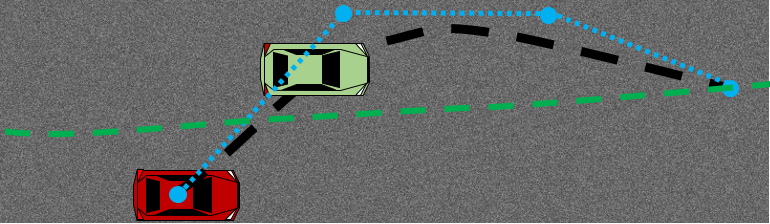}
    \caption{One typical scenario where the reference Bezier-curve has a conflict with the surrounding vehicle when approaching with a large lateral difference.}
    \label{fig:bezier_corner_case}
\end{figure}

Bezier-curves are widely used in path planning algorithms in autonomous driving research \cite{han2010bezier, chen2013lane, qian2016motion} because it is easy to tune and formulate. 
Third-order Bezier-curves are used in this work.
Each Bezier-curve is interpolated from four control points, including shared start and end points with two additional intermediate points, shown in Fig. \ref{fig:control-point}.
Specifically, the start point for the Bezier curve is the ego vehicle's current position and end point is on the time-optimal trajectory generated from learning-based MPC planner.
The selection of end point makes vehicle's state as close as possible to the time-optimal trajectory after overtake behavior.
To make all curves smoother and have no or fewer conflicts with surrounding vehicles, the other two control points will be between the track's boundary and vehicle for Areas 1 and $n$+1, 
shown Fig.~\ref{subfig:control-point-boundary}, 
or between two adjacent vehicles, shown in Fig.~\ref{subfig:control-point-vehicle}.
These two intermediate control points will have the same lateral deviation from the center line.
The key advantage of our selection of control points is that the interpolated geometric curve won't cross the connected lines between control points with its convexity, shown in Fig. \ref{fig:control-point}.
This property makes our reference paths collision-free with respect to surrounding vehicles in most cases, which speed up the computational time of the trajectory generation at each area.

\begin{remark}
In order to avoid mixed-integer and nonlinear optimizations to speed up computation, the planners are separated into (a) the overtaking planner that generates different homotopic reference paths; and (b) trajectory generator that generates dynamically-feasible and collision-free trajectories (in parallel) based on the generated homotopic reference paths.
The overtaking planner generates potential reference geometric paths for obstacle avoidance by choosing control points of the Bezier curves in different homotopic regions. However, this does not guarantee the reference path is collision free.  For instance, when the ego vehicle is approaching other surrounding vehicles with a big lateral difference (see Fig.~\ref{fig:bezier_corner_case}), the Bezier-curve path used in the planner might not be collision-free with other surrounding vehicles %
(while the Bezier control points are.) The parallelized trajectory optimizer then runs multiple trajectory optimizations in parallel for each homotopy class, warm starting from the corresponding Bezier curve. One of the generated dynamically-feasible and collision-free trajectories is then chosen to be sent to the overtaking controller.
\end{remark}

The details of the optimization formulation for trajectory generation will be illustrated in the next section.

\subsection{Trajectory Generation}
\label{subsec:trajectory-planning}
After illustrating the planing strategy, this subsection will show the details about the optimization problem used for trajectory generation for each potential area with different homotopic paths that the ego vehicle can use to overtake the surrounding vehicles.

The optimization problem is formulated as follows,
\begin{subequations}
\label{eq:traj_optimization}
\begin{align}
& \argmin_{x_{t:t{+}N_p|t}, u_{t:t{+}N_p{-}1|t}} p(x_{t{+}N|t}){+}\sum\limits_{k=0}^{N_p-1}q(x_{t{+}k|t})  
 \notag \\
& \quad {+}\sum\limits_{k=1}^{N_p-1}r(x_{t{+}k|t}, u_{t{+}k|t}, x_{t{+}k{-}1|t}, u_{t{+}k{-}1|t}) \label{eq:cost_planner} \\ 
\text{s.t.} \ & x_{t+k+1|t} = A x_{t+k|t} + B u_{t+k|t}, k= 0,...,N_p{-}1 \label{eq:dynamics_update}  \\ 
& x_{t+k+1|t} \in \mathcal{X}, u_{t+k|t} \in \mathcal{U}, k= 0,...,N_p{-}1 \label{eq:state_input_constraint} \\
& x_{t|t} = x_{t}, ~~~~~ \label{eq:initial_state} \\ 
& g(x_{t+k+1|t}) \geq d + \epsilon, k= 0,...,N_p{-}1 \label{eq:safety_constraint}
\end{align}
\end{subequations}
where \eqref{eq:dynamics_update}, \eqref{eq:state_input_constraint}, \eqref{eq:initial_state} are constraints for system dynamics, state/input bounds and initial condition.
The system dynamics constraint describes the affine linearized model described in \eqref{eq:time_invariant_model}.
The cost function \eqref{eq:cost_planner} is composed with three parts along the horizon length $N_p$, the terminal cost $p(x_{t+N|t})$, the stage cost $q(x_{t+k|t})$ and the state/input changing rate cost $ r(x_{t+k|t}, u_{t+k|t})$.
The construction of cost function and constraints in the optimization will be presented in details in the following subsections.

\subsubsection{Terminal Cost}
Terminal cost is about the ego vehicle's traveling distance along the track during overtaking process.
\begin{equation}
\begin{aligned}
\label{eq:terminal_cost}
     p(x_{t+N|t}) = K_d(s_{c_{t+N|t}}-s_{c_{t}})
\end{aligned}
\end{equation}
This compares the open-loop predicted traveling distance at the $N$-th step $s_{c_{t+N|t}}$ with the ego vehicle's current traveling distance $s_{c_t}$. 
This works as the cost metric for timing during the overtaking process.

\subsubsection{Stage Cost}
The stage cost introduces the lateral position differences between the open-loop predicted trajectory and other two paths along the horizon.
\small{
\begin{equation}
\begin{aligned}
\label{eq:stage_cost}
q(x_{t+k|t}){=}||x_{t+k|t}{-}x_{R}(s_{c_k})||^2_{Q_1}{+}||x_{t+k|t}{-}x_{T}(s_{c_k})||^2_{Q_2}
\end{aligned}
\end{equation}}
\normalsize
$x_{R}$ and $x_{T}$ are the reference path and time-optimal trajectory in Frenet coordinates.
The time-optimal trajectory is generated by the learning-based MPC trajectory planner used on a track without other agents, discussed in Sec. \ref{subsec:lmpc}.
$s_{c_k}$ is an initial guess for the traveling distance at the $k$-th step, which is equal to $s_{c_k} = s_{c_t} + v_{x_t} k \Delta_t$, where a constant longitudinal speed is assumed along the prediction horizon.
\subsubsection{State/Input Changing Rate Cost}
To make the predicted trajectory smoother, the state/input changing rate cost $r(x_{t+k|t}, u_{t+k|t})$ is formulated as follow:
\begin{equation}
\begin{aligned}
\label{eq:state_input_changing_cost}
& r(x_{t+k|t}, u_{t+k|t}, x_{t{+}k{-}1|t}, u_{t{+}k{-}1|t}) \\
= & ||x_{t+k|t} - x_{t+k-1|t}||^2_{R_1} + ||u_{t+k|t} - u_{t+k-1|t}||^2_{ R_2}
\end{aligned}
\end{equation}

\subsubsection{Obstacle Avoidance Constraint}
In order to generate a collision-free trajectory, collision avoidance constraint \eqref{eq:safety_constraint} is added in the optimization problem. To reduce computational complexity, only linear lateral position constraint will be added when the ego vehicle overlapes with other vehicles longitudinally. $|s_c(t) + v_x(t) k \Delta t - s_{c,i}(t+k)|<l+\epsilon$ will be used to check if the ego vehicle is overlapping with other vehicle longitudinally along the horizon. In \eqref{eq:safety_constraint}, $g(x) =  |e_{y,i} - e_y| $ shows the lateral position difference, $l$ and $d$ are the vehicle's length and width, $\epsilon$ is a safe margin.

After parallel computation, the optimized trajectory $x^{*}_{t:t+N|t}$ with the minimum cost $J_{s}(x_t)$ discussed in \eqref{eq:cost-selection} will be selected from $n+1$ groups of optimization problems.
It will be tracked by the MPC controller introduced in \ref{sub_sec:overtaking_controller}.

\subsection{Overtaking Controller}\label{sub_sec:overtaking_controller}
After introducing the algorithm for trajectory generation, a low-level tracking controller with model predictive control used for overtaking will be discussed in this part.
The constrained optimization problem is described as follows:
\begin{subequations}
\label{eq:mpc-cbf}
\begin{align}
     \argmin_{\tilde{u}_{t:t{+}N_c{-}1|t}, \omega_{1:N_c{-}1}} &\sum_{k=0}^{N_c-1} \tilde{q}(\tilde{x}_{t+k|t},\tilde{u}_{t+k|t}) + p_{\omega} (1-\omega_{k})^2 \label{eq:mpc-cbf-cost}\\
    \text{s.t.} \ 
    \tilde{x}_{t+k+1|t} = & A_{t+k|t} \tilde{x}_{t+k|t} + B_{t+k|t} \tilde{u}_{t+k|t}\label{eq:mpc-cbf-dynamics} 
    \\ +C_{t+k|t}&, \ k= 0,...,N_c{-}1 \nonumber \\ 
    \tilde{x}_{t+k+1|t} \in & \mathcal{X}, \tilde{u}_{t+k|t} \in \mathcal{U}, \ k = 0,...,N_c{-}1 \label{eq:mpc-cbf-state-input-constraint}\\
    \tilde{x}_{t|t} =& \tilde{x}_t, \label{eq:mpc-cbf-initial-condition}\\
    h (\tilde{x}_{t+k+1|t}) \geq & \gamma \omega_{k} h(\tilde{x}_{t+k|t}), \ k = 0,...,N_c{-}1 \label{eq:mpc-cbf-cbf}
\end{align}
where \end{subequations} \eqref{eq:mpc-cbf-dynamics}, \eqref{eq:mpc-cbf-state-input-constraint}, \eqref{eq:mpc-cbf-initial-condition} describe the constraints for system dynamics \eqref{eq:time_varying_model}, input/state bounds and initial conditions, respectively.
The $\tilde{q}(\tilde{x}_{t+k|t}, \tilde{u}_{t+k|t})= ||\tilde{x}_{t+k|t} - x^{*}_{t+k|t}||^2_{\tilde{Q}_1} + ||\tilde{u}_{t+k|t}||^2_{\tilde{Q}_2}$ represents the stage cost, which tracks the desired trajectory $x^{*}_{t:t+N|t}$ optimized by the trajectory planner.
Equation \eqref{eq:mpc-cbf-cbf} with $0\leq\gamma<1$ represents discrete-time control barrier function constraints~\cite{zeng2021enhancing} with relaxation ratio $\omega_k$ for feasibility~\cite{zeng2021pointwise}, which could guarantee the system's safety by guaranteeing $h(\tilde{x}_{t+k|t}) > 0 $ along the horizon with forward invariance. In this project, $h(\tilde{x}_{t+k|t}) = (\tilde{s}_{c,i} - \tilde{s}_c)^2 + (\tilde{e}_{y,i} - \tilde{e}_y)^2 -l^2 - d^2 $ is used to represent the distance between the ego vehicle and other vehicles.
The optimization \eqref{eq:mpc-cbf} allows us to find the optimal control $u^*_t = \tilde{u}_{t|t}$ in a manner similar to MPC.

\begin{remark}
Notice that \eqref{eq:traj_optimization} uses a distance constraint for obstacle avoidance while \eqref{eq:mpc-cbf} uses control barrier functions. The reason why we don't combine these optimizations into one under dynamics \eqref{eq:mpc-cbf-dynamics} is that this cascaded approach for trajectory generation and control is shown to be more computationally efficient and less likely to generate deadlock behavior.
\end{remark}

%% file: sections/results.tex
\section{Results}
\label{sec:Results}
Having illustrated our autonomous racing strategy in the previous section, we now show the performance of proposed algorithm. The setup and results of numerical simulations will be presented in the following part.
\subsection{Simulation Setup}
In all simulations, all vehicles are 1:10 scale RC cars with a length of 0.4m and a width of 0.2m. Other vehicles drive along trajectories with fixed lateral deviation from track's center line.
The track's width is set to 2 m. 
The horizon lengths for trajectory planner and controller are $N_p = 12$, $N_c = 10$ and shared discretization time $\Delta t = 0.1s$.
In our custom-designed simulator, both state and input noises are considered. Surrounding vehicles' speed and position information is used for trajectory generation. However, no interaction between the ego vehicle and other surrounding vehicles is included.

The optimization problems \eqref{eq:traj_optimization} and \eqref{eq:mpc-cbf} are implemented in Python with CasADi~\cite{andersson2019casadi} used as modeling language, are solved with IPOPT~\cite{biegler2009large} on Ubuntu 18.04 on a laptop with a CPU i7-9850 processor at a 2.6Ghz clock rate.

\subsection{Racing With Other Vehicles}
\begin{figure*}
    \centering
    \begin{subfigure}[t]{0.48\linewidth}
        \centering
        \includegraphics[width = 0.99\linewidth]{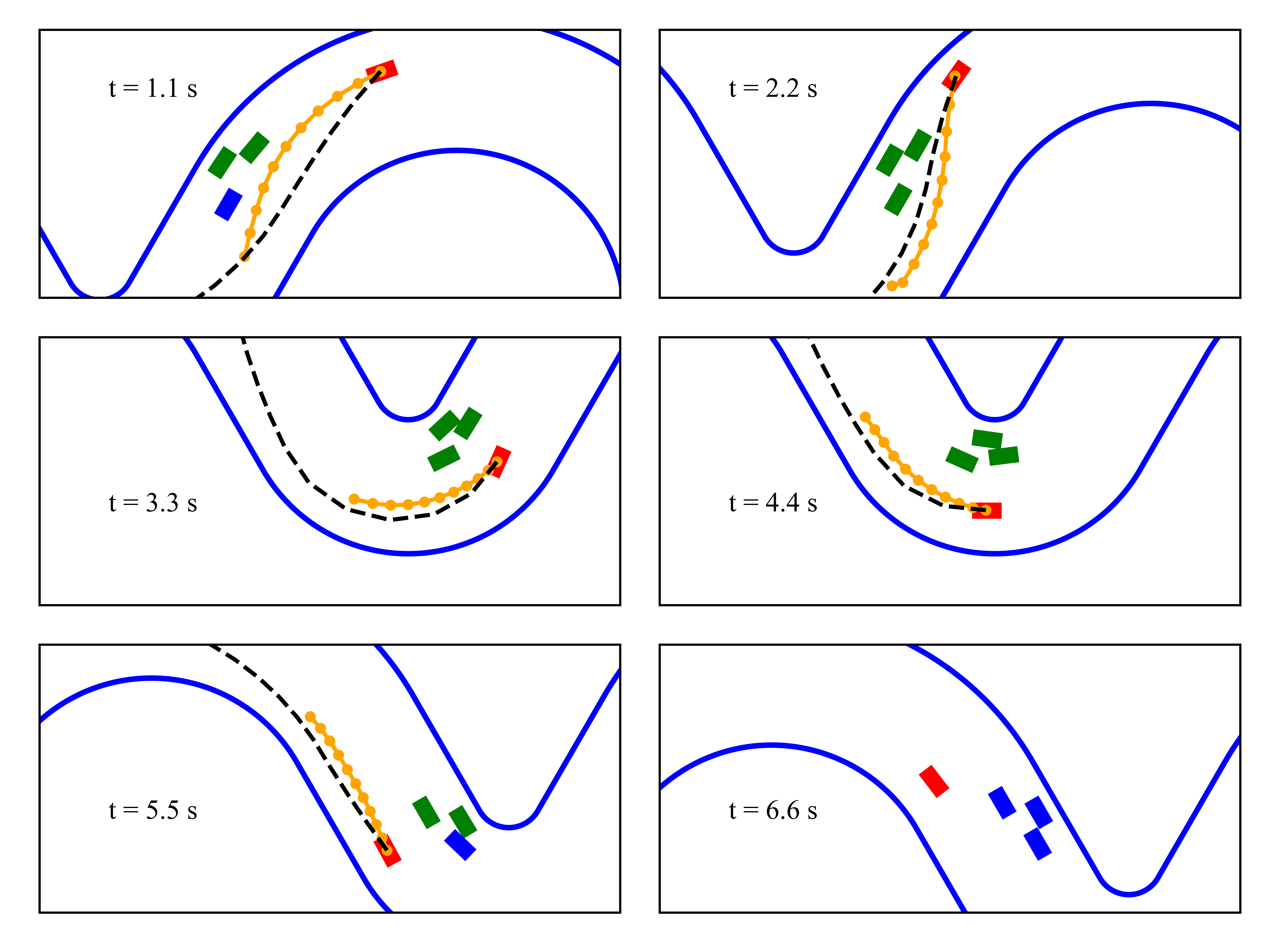}
        \caption{
        Overtaking snapshots on curvy track segment.
        Other surrounding vehicles move at 1.3 m/s, 1.3 m/s and 1.35 m/s.}
        \label{subfig:overtake-example-1}
    \end{subfigure}
    \begin{subfigure}[t]{0.48\linewidth}
        \centering
        \includegraphics[width = 0.99\linewidth]{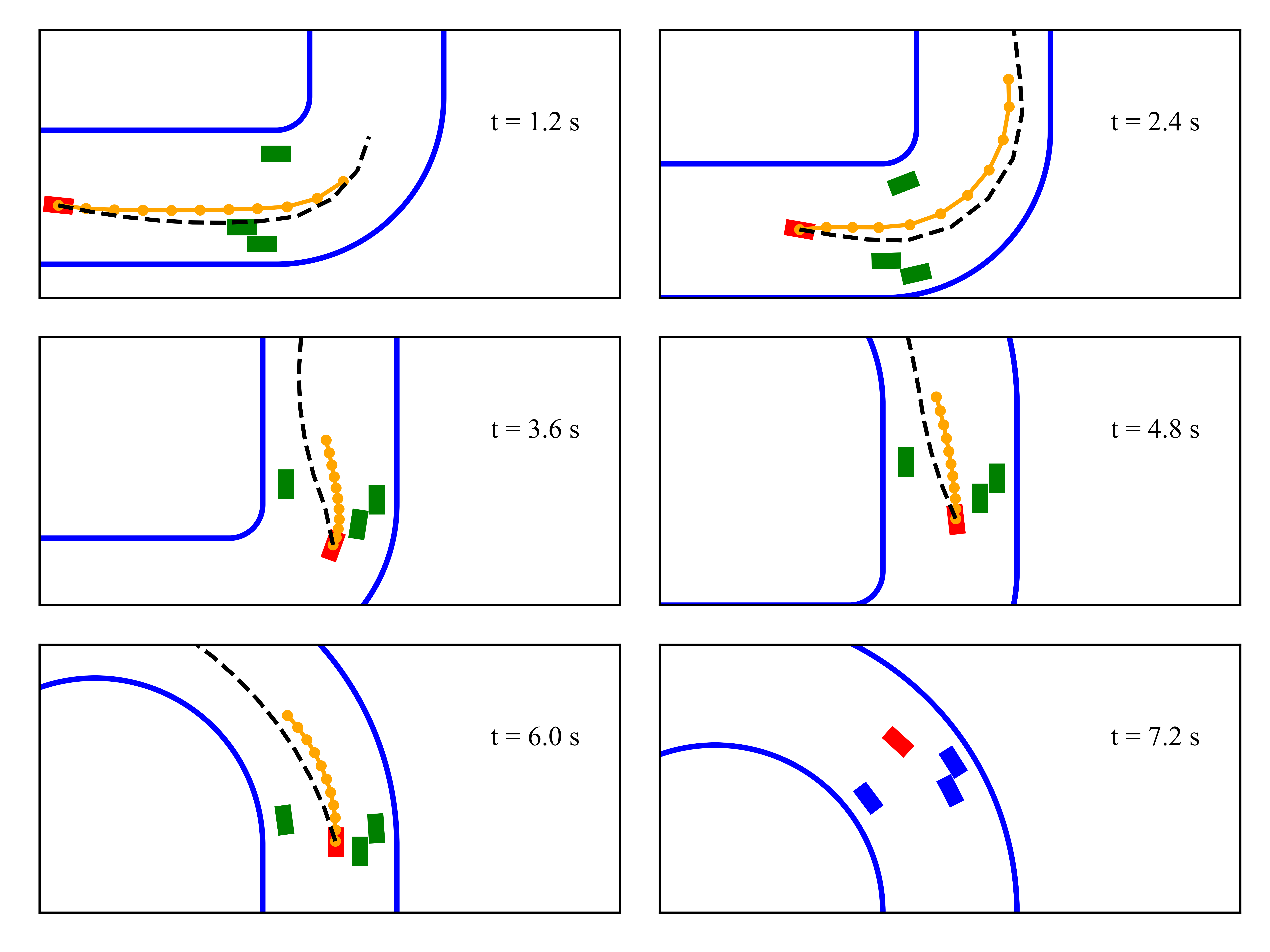}
        \caption{
        Overtaking snapshots on straight track segment.
        Other surrounding vehicles all move at 1.2 m/s.
        }
        \label{subfig:overtake-example-2}
    \end{subfigure}
    \caption{Snapshots from simulation of the overtaking behavior. 
    The red vehicle is the ego vehicle, the orange line shows the optimized trajectory from the planning strategy, and the dashed black line is the reference Bezier-curve used for generating the selected trajectory.
    Other vehicles are marked in blue with those in the ego vehicle's range of overtaking marked in green. The solid blue line is the track's boundary.}
    \label{fig:overtake-snapshots}
\end{figure*}

Snapshots shown in Fig.~\ref{fig:overtake-snapshots} illustrate examples of the overtaking behavior in both straight and curvy track segments when competing with other vehicles.
When the ego vehicle competes with three surrounding vehicles, it could overtake them on one side of all vehicles (Fig.~\ref{subfig:overtake-example-1}) or between them (Fig.~\ref{subfig:overtake-example-2}).
The animations of more challenging overtaking behavior can be found at \url{https://youtu.be/1zTXfzdQ8w4}.
As shown in TABLE~\ref{tab:work_summary}, the proposed racing planner could update at 25 Hz and could help the ego vehicle overtake multiple moving vehicles.
By switching to a trajectory planning based on learning-based MPC, the ego vehicle is able to reach its speed and steering limit when there are no surrounding vehicles.

\begin{table}[!htp]
    \centering
    \begin{tabular}{|c|c|c|c|c|}
    \hline
    Speed Range [m/s] & 0 - 0.4 & 0.4 - 0.8 & 0.8 - 1.2 & 1.2 - 1.6 \\ \hline \hline
    mean [s] & 1.613 & 2.312 & 3.857 & 13.095 \\ \hline
    min [s] & 0.8 & 1.2 & 1.8 & 3.5  \\ \hline
    max [s] & 3.6 & 5.2 & 21.6 & 36.1 \\ \hline
    \end{tabular}
    \caption{Time taken to overtake the leading vehicle travelling at different speeds. For each group of speed range of the leading vehicles, 100 cases were simulated. The ego vehicle starts from the track's origin. One other vehicle starts from a random position in the range of $ 10 m \leq s_{c,i} \leq 30 m$. The mean, min and max values show the average overtaking time, minimum overtaking time and maximum overtaking time for the corresponding group. In general, it takes more time to overtake faster moving vehicles on this track since they spend lesser time on the straight segments.}
    \label{tab:overtaking-time}
\end{table}

To better analyze the performance and limitations of our autonomous racing strategy in different scenarios, random tests are introduced under two groups.
The first group of simulation aims to show the overtaking time for passing one leading vehicle with different speeds, and statistical results are summarized in TABLE \ref{tab:overtaking-time}.
We can observe that when the surrounding vehicles' speed reaches between 1.2m/s and 1.6m/s, much more time is needed for the ego vehicle to overtake the leading vehicle.
This is because as the leading vehicle's speed increases, less space becomes available for the ego vehicle to drive safely.
Especially in a curve, the ego vehicle's speed limit decreases when less space can be used to make a turn.
Since more than half of our track is with curves, the ego vehicle needs to wait for a straight segment to accelerate to pass the leading vehicle.

\begin{table}[!htp]
    \centering
    \begin{tabular}{|c|c|c|c|c|}
    \hline
    Speed Range [m/s] & 0 - 0.4 & 0.4 - 0.8 & 0.8 - 1.2 & 1.2 - 1.6 \\ \hline \hline
    Single & 100 $\%$ & 100 $\%$ & 96 $\%$ & 84 $\%$ \\ \hline
    Two & 100 $\%$ & 100 $\%$ & 98 $\%$ & 66 $\%$ \\ \hline
    Three & 100 $\%$ & 98 $\%$ & 84 $\%$ & 36 $\%$ \\ \hline
    \end{tabular}
    \caption{Overtaking success rate for the ego vehicle after one lap.
    For each group of speed range of the leading vehicles, 100 cases were simulated.
    The ego vehicle starts from the track's origin. Other vehicles start from a random position in the range of $ 5 m \leq s_{c,i} \leq 15 m$. One to three leading cars were simulated.}
    \label{tab:overtaking-success-rate}
\end{table}

The second group of simulation shows the proposed racing strategy's success rate to overtake multiple leading vehicles in one lap, and statistical results are summarized in TABLE \ref{tab:overtaking-success-rate}.
We can find that when more than one surrounding vehicle exists, much more space would be occupied by other vehicles. As a result, the ego vehicle might not have enough space to accelerate to high speed to pass surrounding vehicles.
Although in these cases, the ego vehicle can not overtake all surrounding vehicles after one lap, our proposed racing strategy can still guarantee the ego vehicle's safety along the track.

During our simulation, the mean solver time for our planner for single, two or three surrounding vehicles is 39.21ms, 39.41ms and 40.23ms.
We also notice that when the number of surrounding vehicles is larger than three, the steady complexity still holds but the track becomes too crowded for the ego vehicle to achieve high success rate of overtake maneuver. 
This validates the steady low computational complexity of proposed planning strategy thanks to the parallel computation for multiple trajectory optimizations.

Moreover, the optimized multiple trajectory candidates offer more choices for overtaking, which helps the ego vehicle avoid deadlock.

\begin{figure}[th]
    \setlength{\abovecaptionskip}{0.5cm}
    \setlength{\belowcaptionskip}{-0.4cm}
    \centering
    \includegraphics[width=1.0\linewidth]{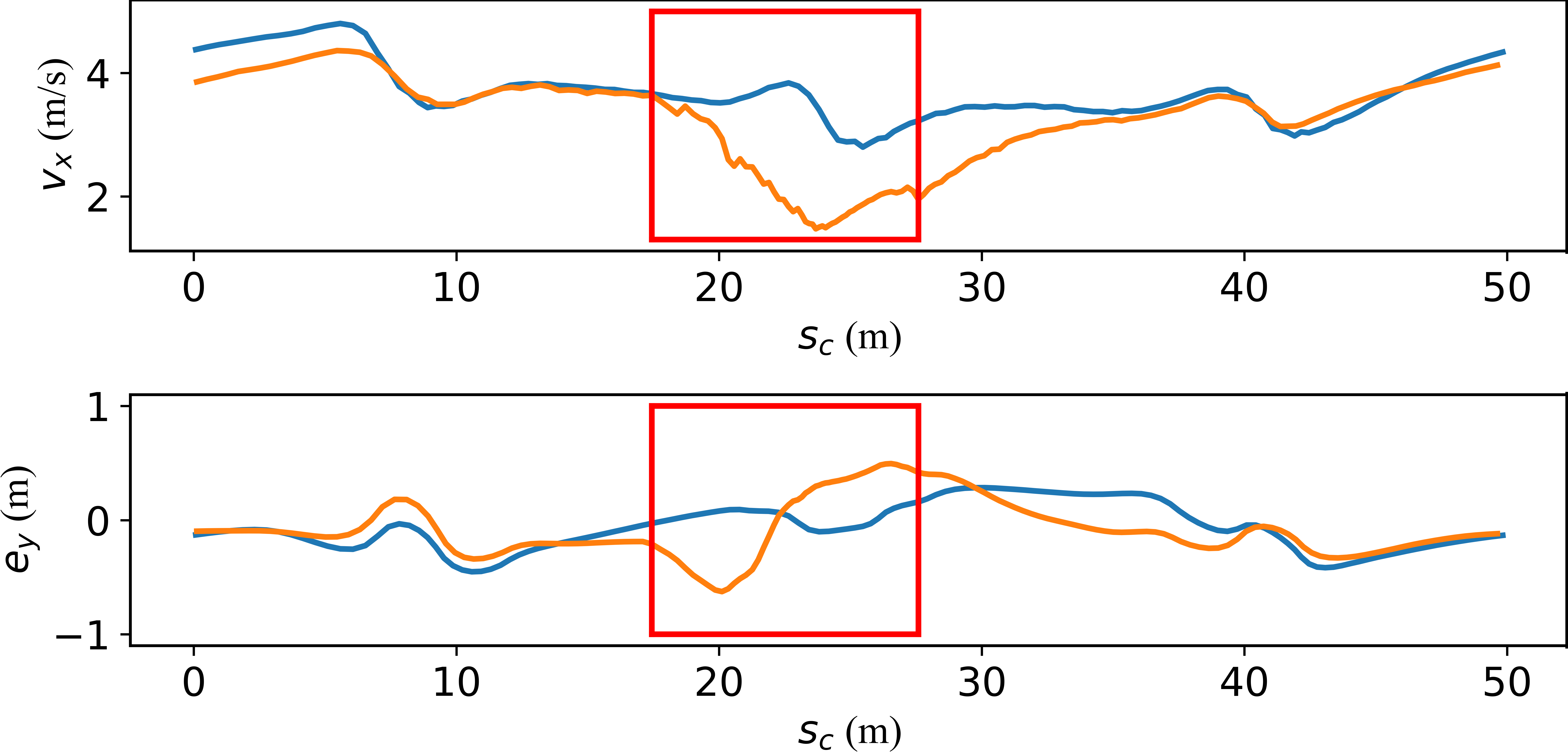}
    \caption{Speed and lateral deviation from the track's center line. We illustrate our proposed strategy with three competing vehicles (orange line) as well as the learning-based MPC strategy without any competing vehicles (blue line). The highlighted red region shows the session for overtaking.}
    \label{fig:speed-comparasion}
\end{figure}

\subsection{Racing Without Other Vehicles}
As discussed in Sec. \ref{subsec:racing-strategy}, when there are no other surrounding vehicles, the ego vehicle adopts the learning-based MPC formulation for trajectory generation and control.
In this paper, the learning-based MPC uses historical data from two previous laps (implying $k = 2$ in Fig. \ref{fig:lmpc_illustration})
and the initial data are calculated offline before the racing tasks. For the same setup as shown in Fig.~\ref{subfig:overtake-example-1}, the ego vehicle's speed and deviation from track's center line along the track is shown with learning-based MPC's profile in Fig.~\ref{fig:speed-comparasion}.
The overtake maneuver happens in a hairpin curve and the curve's apex is occupied by other moving vehicles, resulting in less space being available for the ego vehicle and thus causing it to slow down to avoid a potential collision. After it passes all surrounding vehicles, the ego vehicle goes back to drive at its speed and steering limit to achieve time-optimal behavior.

%% file: sections/conclusion.tex
\section{Conclusion} \label{sec:Conclusion}
In this paper, we have presented an autonomous racing strategy that enables an ego vehicle to enhance its lap timing performance while overtaking other moving vehicles. 
We have verified the performance of our proposed algorithm through numerical simulation, where several surrounding vehicles are simulated to start from random positions with random speeds on a track.
Moreover, interaction between the ego vehicle and other surrounding vehicles will be considered in the future work. 
For instance, autonomous racing strategies such as blocking cars from overtaking are envisaged for the future.